\documentclass[runningheads]{llncs}
\usepackage{subcaption}
\usepackage{graphicx}

\usepackage{tikz}
\usepackage{comment}
\usepackage{amsmath,amssymb} 
\usepackage{color}

\usepackage[accsupp]{axessibility}  

\usepackage{booktabs}
\usepackage{multirow}
\usepackage{wrapfig}

\usepackage[pagebackref=true,breaklinks=true,colorlinks,bookmarks=false]{hyperref}

\newcommand{\eg}{\textit{e.g.},\ }
\newcommand{\etal}{et al.\ }
\newcommand{\ie}{\textit{i.e.},\ }
\newcommand{\wrt}{w.r.t.\ }

\newcommand{\myparagraph}[1]{\noindent\textbf{#1}:~}

\definecolor{goldenrod}{rgb}{0.85, 0.65, 0.13}
\definecolor{carolinablue}{rgb}{0.6, 0.73, 0.89}
\definecolor{darkmidnightblue}{rgb}{0.0, 0.2, 0.4}
\definecolor{asparagus}{rgb}{0.53, 0.66, 0.42}

\newcommand{\sectionvspace}{}
\newcommand{\subsectionvspace}{}
\newcommand{\captionvspace}{}

\begin{document}
\pagestyle{headings}
\mainmatter
\def\ECCVSubNumber{4029}  

\title{Map-free Visual Relocalization:\\Metric Pose Relative to a Single Image}

\titlerunning{Map-free Visual Relocalization}

\author{Eduardo Arnold$^{1,2}$\thanks{Work done during internship at Niantic} \and
Jamie Wynn$^1$ \and
Sara Vicente$^1$ \and \\
Guillermo Garcia-Hernando$^1$ \and
{\'{A}}ron Monszpart$^1$ \and
Victor Adrian Prisacariu$^{1,3}$ \and \\
Daniyar Turmukhambetov$^1$ \and
Eric Brachmann$^1$}
\authorrunning{E. Arnold et al.}
%
\institute{$^1$Niantic \hspace{12pt} $^2$University of Warwick \hspace{12pt}  $^3$University of Oxford\\
\vspace{4pt}
\href{http://github.com/nianticlabs/map-free-reloc}{\nolinkurl{github.com/nianticlabs/map-free-reloc}}}

\maketitle

\begin{abstract}

Can we relocalize in a scene represented by a single reference image? 
Standard visual relocalization requires hundreds of images and scale calibration to build a scene-specific 3D map. 
In contrast, we propose \mbox{\emph{Map-free Relocalization}}, \ie using only one photo of a scene to enable instant, metric scaled relocalization.
Existing datasets are not suitable to benchmark map-free relocalization, due to their focus on large 
scenes or their limited variability. 
Thus, we have constructed a new dataset of 655 small places of interest, such as sculptures, murals and fountains, collected worldwide. 
Each place comes with a reference image to serve as a relocalization anchor, and dozens of query images with known, metric camera poses. 
The dataset features changing conditions, stark viewpoint changes, high variability across places, and queries with low to no visual overlap with the reference image.
We identify two viable families of existing methods to provide baseline results:
relative pose regression, and feature matching combined with single-image depth prediction. 
While these methods show reasonable performance on some favorable scenes in our dataset, map-free relocalization proves to be a challenge that requires new, innovative solutions.
\end{abstract}

\section{Introduction}
\label{sec:intro}
\sectionvspace

\begin{figure}[t]
  \centering
  \includegraphics[width=1.0\linewidth]{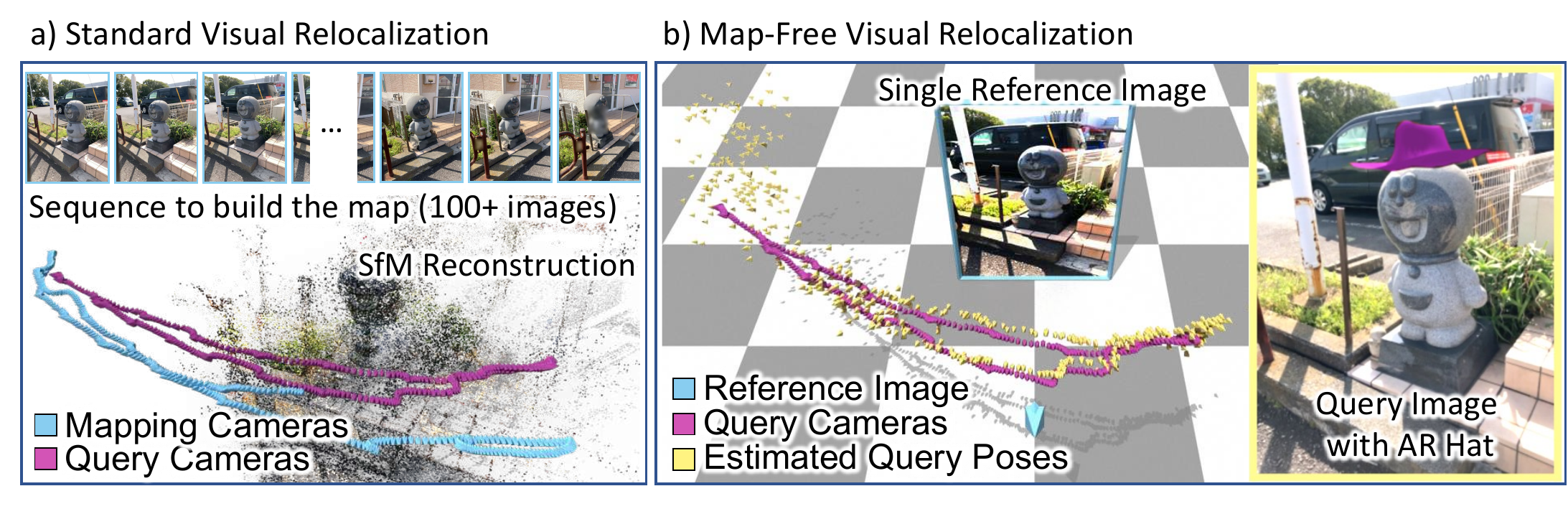}

   \caption{Standard relocalization methods build a scene representation from hundreds of mapping images (a). For map-free relocalization (b), only a single photo (cyan) of the scene is available to relocalize queries. We show ground truth poses (purple) and estimated poses (yellow) in (b), and we use one estimate to render a virtual hat on the statue. We achieve these results with SuperGlue~\cite{sarlin2020superglue} feature matching and DPT~\cite{ranftl2021DPT} depth estimation.
   }
   \label{fig:teaser}
   \captionvspace
\end{figure}

Given not more than a single photograph we can imagine what a depicted place looks like, and where we, looking through the lens, would be standing relative to that place. 
Visual relocalization mimics the human capability to estimate a camera's position and orientation from a single query image.
It is a well-researched task that enables exciting applications in augmented reality (AR) and robotic navigation. 
State-of-the-art relocalization methods surpass human rule-of-thumb estimates by a noticeable margin~\cite{brachmann2017dsac,kapture2020,sarlin2019coarse,sarlin2020superglue,sattler2011fast,sattler2018benchmarking}, allowing centimeter accurate predictions of a camera's pose. 
But this capability comes with a price: 
each scene has to be carefully pre-scanned and reconstructed. 
First, images need to be gathered from hundreds of distinct viewpoints, ideally spanning different times of day and even seasons. 
Then, the 3D orientation and position of these images needs to be estimated, \eg by running structure-from-motion (SfM)~\cite{schoenberger2016sfm,snavely2006photo,wu2011visualsfm,wu2013towards} or simultaneous-localization-and-mapping (SLAM)~\cite{dai2017bundlefusion,newcombe2011kinectfusion} software. 
Oftentimes, accurate multi-camera calibration, alignment against LiDAR scans, high-definition maps or inertial sensor measurements are needed to recover poses in metric units, \eg~\cite{sattler2018benchmarking,sattler2012image}.
Finally, images and their camera poses are fed to a relocalization pipeline.
For traditional structure-based systems \cite{kapture2020,sarlin2019coarse,sattler2011fast,Sattler2017AS}, the final scene representation consists of a point cloud triangulated from feature correspondences, and associated feature descriptors, see \mbox{Fig.~\ref{fig:teaser}a)}.

The requirement for systematic pre-scanning and mapping restricts how visual relocalization can be used. 
For example, AR immersion might break if a user has to record an entire image sequence of an unseen environment first, gathering sufficient parallax by walking sideways, all in a potentially busy public space.
Furthermore, depending on the relocalization system, the user then has to wait minutes or hours until the scene representation is built. 
We propose a new flavour of relocalization, termed \emph{Map-free Relocalization}.
We ask whether the mapping requirement can be relaxed to the point where a single reference image is enough to relocalize new queries in a metric coordinate system. 
Map-free relocalization enables instant AR capabilities at new locations: User A points their camera at a structure, takes a photo, and any user B can instantly relocalize \wrt user A.
Map-free relocalization constitutes a systematic, task-oriented benchmark for two-frame relative pose estimation, namely between the reference image and a query image, see Fig.~\ref{fig:teaser} b). 

Relocalization by relative pose estimation is not new. 
For example, neural networks have been trained to regress metric relative poses directly from two images~\cite{balntas2018relocnet,melekhov2017relative}.
Thus far, such systems have been evaluated on standard relocalization benchmarks where structure-based methods rule supreme, to the extent where the accuracy of the ground truth is challenged~\cite{brachmann2021limits}. 
We argue that we should strive towards enabling new capabilities that traditional structure-based methods cannot provide.
Based on a single photo, a scene cannot be reconstructed by SfM or SLAM.
And while feature matching still allows to estimate the relative pose between two images, the reference and the query, there is no notion of absolute scale~\cite{hartley2003multiple}.
To recover a \emph{metric} estimate, some heuristic or world knowledge has to be applied to resolve the scale ambiguity which we see as the key problem.

Next to pose regression networks, that predict metric poses by means of supervised learning, we recognize a second family of methods as suitable for map-free relocalization.
We show that a combination of deep feature matching~\cite{sarlin2020superglue,sun2021loftr} and deep single-image depth prediction~\cite{ranftl2021DPT,liu2019planercnn} currently achieves highest relative pose accuracy. 
To the best of our knowledge, this variant of relative pose estimation has not gained attention in relocalization literature thus far. 

While we provide evidence that existing methods can solve map-free relocalization with acceptable precision, such results are restricted to a narrow window of situations.
To stimulate further research in map-free relocalization, we present a new benchmark and dataset.
We have gathered images of 655 places of interest worldwide where each place can be represented well by a single reference image. All frames in each place of interest have metric ground truth poses.
There are 522,921 frames for training, 36,998 query frames across 65 places for validation, and 14,778 query frames (subsampled from 73,902 frames) across 130 places in the test set.
Following best practice in machine learning, we provide a public validation set while keeping the test ground truth private, accessed through an online evaluation service.
This dataset can serve as a test bed for advances in relative pose estimation and associated sub-problems such as wide-baseline feature matching, robust estimation and single-image depth prediction.

\noindent We summarize our \textbf{contributions} as follows:
{
\begin{itemize}
    \item Map-free relocalization, a new flavor of visual relocalization that dispenses with the need for creating explicit maps from extensive scans of a new environment. A single reference image is enough to enable relocalization.
    \item A dataset that provides reference and query images of over 600 places of interest worldwide, annotated with ground truth poses. The dataset includes challenges such as changing conditions, stark viewpoint changes, high variability across places, and queries with low to no visual overlap with the reference image.
    \item Baseline results for map-free relocalization using relative pose regression methods, and feature matching on top of single image-depth prediction. We expose the primary problems of current approaches to guide further research.
    \item Additional experiments and ablation studies on ScanNet and 7Scenes datasets, allowing comparisons to related, previous research on relative pose estimation and visual relocalization.
\end{itemize}
}

\section{Related Work}
\sectionvspace

\myparagraph{Scene Representations in Visual Relocalization}
In the introduction, we have discussed traditional structure-based relocalizers that represent a scene by an explicit SfM or SLAM reconstruction. 
As an alternative, recent learning-based relocalizers encode the scene \emph{implicitly} in the weights of their neural networks by training on posed mapping images.
This is true for both scene coordinate regression~\cite{shotton2013scene,brachmann2017dsac,Brachmann2018dsacpp,Brachmann2019ESAC,li2020hierarchical,Brachmann2021dsacstar} and absolute pose regression (APR)~\cite{kendall2015posenet,Kendall2017GeometricLF,Walch2017lstm,Brahmbhatt2018mapnet,Shavit2021MStransformer}.
More related to our map-free scenario, some relative pose regression (RPR) methods avoid training scene specific networks~\cite{balntas2018relocnet,turkoglu2021visual,WinkelbauerICRA21}.
Given a query, they use image retrieval~\cite{torii2015denseVLAD,netvlad,boxes,kapture2020,radenovic2016cnn,pion2020benchmarking,sattler2012image} to look up the closest database image and its pose.
A generic relative pose regression network estimates the pose between query and database images to obtain the absolute pose of the query.
RPR methods claim to avoid creating costly scene-specific representations but ultimately these works do not discuss how posed database images would be obtained without running SfM or SLAM.  
ExReNet\cite{WinkelbauerICRA21}, a recent RPR method, shows that the database of posed images can be extremely sparse, keeping as little as four strategically placed reference images to cover an indoor room. 
Although only a few images make up the final representation, continuous pose tracking is required when recording them.
In contrast, map-free relocalization means keeping only a single image to represent a scene without any need for pose tracking or pose reconstruction.
The reference image has the identity pose.

\myparagraph{Relative Pose by Matching Features}
The pose between two images with known intrinsics can be recovered by decomposing the essential matrix \cite{hartley2003multiple}.
This yields the relative rotation, and a \emph{scaleless} translation vector.
The essential matrix is classically estimated by matching local features, followed by robust estimation, such as using a 5-point solver~\cite{nister2004fivept} inside a RANSAC~\cite{fischler1981random} loop.
This basic formula has been improved by learning better features~\cite{sift,revaud2019r2d2,dusmanu2019d2,Tyszkiewicz2020DISK,Bhowmik2020ReFP}, better matching~\cite{sarlin2020superglue,sun2021loftr} and better robust estimators~\cite{raguram2008comparative,yi2018learning,ranftl2018deep,brachmann2019NGransac,barath2019magsac,barath2019magsacplusplus,Sun2020acne}, and progress has been measured in wide-baseline feature matching challenges~\cite{Jin2020} and small overlap regimes.

In the relocalization literature, scaleless pairwise relative poses between the query and multiple reference images have been used to triangulate the scaled, metric pose of a query~\cite{zhang2006image,zhou2020essnet,WinkelbauerICRA21}.
However, for map-free relocalization only two images (reference and query) are available at any time, making query camera pose triangulation impossible.
Instead, we show that estimated depth can be used to resolve the scale ambiguity of poses recovered via feature matching.

\myparagraph{Relative Pose Regression (RPR)}
Deep learning methods that predict the relative pose from two input images bypass explicit estimation of 2D correspondences~\cite{ummenhofer2017demon,melekhov2017relative,en2018rpnet,balntas2018relocnet,WinkelbauerICRA21,abouelnaga2021distillpose}. 
Some methods recover pose up to a scale factor~\cite{melekhov2017relative,WinkelbauerICRA21} and rely on pose triangulation, while others aim to estimate metric relative pose~\cite{balntas2018relocnet,en2018rpnet,abouelnaga2021distillpose}.
Both RelocNet~\cite{balntas2018relocnet} and ExReNet~\cite{WinkelbauerICRA21} show generalization of RPR across datasets by training on data different from the test dataset.

Recently, RPR was applied in scenarios that are challenging for correspondence-based approaches. 
Cai~\etal~\cite{cai2021extreme} focus on estimating the relative rotation between two images in extreme cases, including when there is no overlap between the two images. 
Similarly, the method in~\cite{chen2021wide} estimates scaleless relative pose for pairs of images with very low overlap. 
We take inspiration from the methods above to create baselines and discuss in more detail the different architectures and output parameterizations in Section~\ref{sec:regression_methods}.

\myparagraph{Single-image Depth Prediction}
Advances in deep learning have allowed practical methods for single-image depth estimation, \eg \cite{eigen2014depth,monodepth}. 
There are two versions of the problem: relative and absolute depth prediction. 
Relative, also called scaleless, depth prediction aims at estimating depth maps up to an unknown linear or affine transformation, and can use scaless training data such as SfM reconstructions~\cite{li2018megadepth}, uncalibrated stereo footage~\cite{xian2018monocular,Ranftl2020midas} or monocular videos~\cite{zhou2017unsupervised,monodepth2}.
Absolute depth prediction methods (\eg \cite{ranftl2021DPT,liu2019planercnn,monodepth,depthhints}) aim to predict depth in meters by training or fine-tuning on datasets that have absolute metric depth such as the KITTI~\cite{Geiger2012kitti}, NYUv2~\cite{silberman2012NYUdataset} and ScanNet~\cite{dai2017scannet} datasets. 
Generalizing between domains (\eg driving scenes vs.~indoors) is challenging as collecting metric depth in various conditions can be expensive. 
Moreover, generalization of a single network that is robust to different input image resolutions, aspect ratios and camera focal lengths is also challenging~\cite{facil2019cam}.

Recently, single-image depth prediction was leveraged in some pose estimation problems.
In~\cite{toft2020single}, predicted depth maps are used to rectify planar surfaces before local feature computation for improved relative pose estimation under large viewpoint changes.
However, that work did not use metric depth information to estimate the scale of relative poses. 
Depth prediction was incorporated into monocular SLAM~\cite{tateno2017cnn,tiwari2020pseudo} and Visual Odometry~\cite{yang2018deepVSO,campos2021scale} pipelines to combat scale drift and improve camera pose estimation. 
Predicted depths were used as a soft constraint in multi-image problem, while we use depth estimates to scale relative pose between two images.

\section{Map-free Relocalization}
\sectionvspace

Our aim is to obtain the camera pose of a query image given a single RGB reference image of a scene.
We assume intrinsics of both images are known, as they are generally reported by modern devices.
The absolute pose of a query image $Q$ is parameterized by $R \in SO(3), \, t \in \mathbb{R}^3$, which maps a world point $\mathbf{y}$ to point $\mathbf{x}$ in the camera's local coordinate system as $\mathbf{x} = R \mathbf{y} + t$.
Assuming the global coordinate system is anchored to the reference image, the problem of estimating the absolute pose of the query becomes one of estimating a scaled relative pose between two images.
Next, we discuss different approaches for obtaining a metric relative pose between a pair of RGB images.
The methods are split into two categories: methods based on feature matching with estimated depth, and methods based on direct relative pose regression.

\begin{figure}[t]
  \centering
   \includegraphics[width=0.99\linewidth]{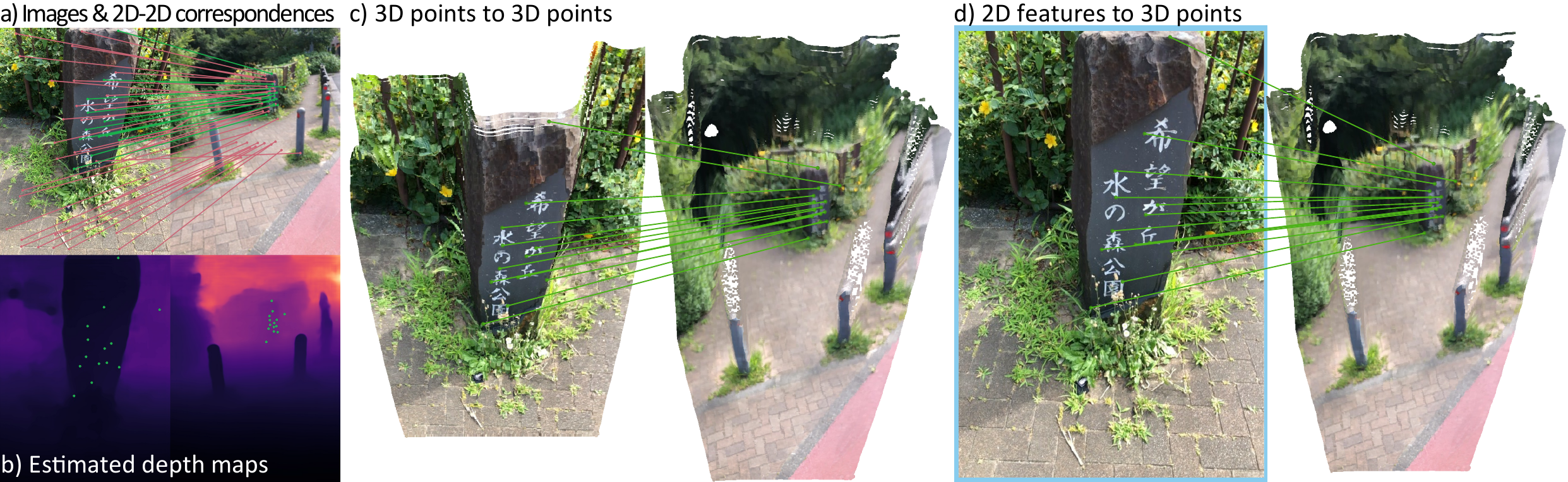}
   \caption{Given the reference and query images, we obtain 2D-2D correspondences using the feature matching method in~\cite{sarlin2020superglue} (a). Inlier correspondences for the robust RANSAC-based essential matrix computation are visualized in green and outlier correspondences in red.
   Estimated monocular depth maps using~\cite{ranftl2021DPT} are shown in (b). 
   The depth maps can be coupled with the 2D-2D correspondences to obtain 3D-3D correspondences (c) or 2D-3D correspondences (d), which are used in the geometric methods discussed in Section~\ref{sec:feature_matching_depth}.
   }
   \label{fig:scaled_poses}
   \captionvspace
\end{figure}

\subsection{Feature Matching and Scale from Estimated Depth}\label{sec:feature_matching_depth}
\subsectionvspace
The relative pose from 2D correspondences is estimated up to scale via the Essential matrix \cite{hartley2003multiple}.
We consider SIFT~\cite{sift} as a traditional baseline as well as more recent learning-based matchers such as  SuperPoint + SuperGlue~\cite{sarlin2020superglue} and LoFTR~\cite{sun2021loftr}.  
To recover the missing scale, we utilize monocular depth estimation.
For indoors, we experimented with DPT~\cite{ranftl2021DPT} fine-tuned on the NYUv2 dataset~\cite{silberman2012NYUdataset} and PlaneRCNN~\cite{liu2019planercnn}, which was trained on ScanNet~\cite{dai2017scannet}. 
For outdoors, we use DPT~\cite{ranftl2021DPT} fine-tuned on KITTI~\cite{Geiger2012kitti}.
Given estimated depth and 2D correspondences we compute scaled relative poses in the following variants. See also Fig.~\ref{fig:scaled_poses} for an illustration.

\myparagraph{(2D-2D) Essential matrix + depth scale (\emph{Ess.Mat. + D.Scale)}}
We compute the Essential matrix using a 5-point solver~\cite{nister2004fivept} with MAGSAC++~\cite{barath2019magsacplusplus} and decompose it into a rotation and a unitary translation vector.
We back-project MAGSAC inlier correspondences to 3D using the estimated depth.
Each 3D-3D correspondence provides one scale estimate for the translation vector, and we select the scale estimate with maximum consensus across correspondences, see the supplemental material for details.

\myparagraph{(2D-3D) Perspective-n-Point (PnP)} 
Using estimated depth, we back-project one of the two images to 3D, giving 2D-3D correspondences.
This allows us to use a PnP solver \cite{gao2003complete} to recover a metric pose. 
We use PnP within RANSAC \cite{fischler1981random} and refine the final estimate using all inliers. 
We use 2D features from the query image and 3D points from the reference image.

\myparagraph{(3D-3D) Procrustes} 
Using estimated depth, we back-project both images to 3D, giving 3D-3D correspondences.
We compute the relative pose using Orthogonal Procrustes~\cite{eggert1997rigidtransforms} inside a RANSAC loop \cite{fischler1981random}.
Optionally, we can refine the relative pose using ICP~\cite{mckay1992icp} on the full 3D point clouds.
This variant performs significantly worse compared to the previous two, so we report its results in the supplemental material.

\begin{figure}[t] 
  \centering
   \includegraphics[width=0.95\linewidth]{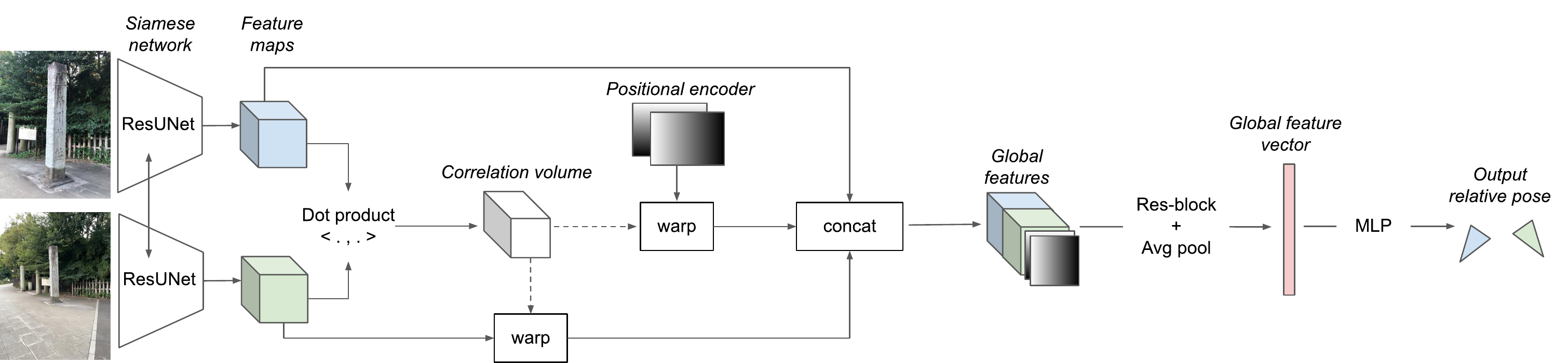}
    \caption{Overview of the network architecture for RPR. We use a Siamese network (ResUNet~\cite{cai2021extreme}) to extract features from the two input images. Following~\cite{cai2021extreme,WinkelbauerICRA21}, we compute a 4D correlation volume to mimic soft feature matching.
    The correlation volume is used to warp the features of the second image and a regular grid of coordinates (positional encoding). These are concatenated channel-wise with the first image's feature map to create the global feature map.
    The global volume is fed to four ResNet blocks followed by global average pooling, resulting in a single 512-dimensional global feature vector. Finally, an MLP generates the output poses. See supplement for details.}
   \label{fig:network_architecture}
   \captionvspace
\end{figure}

\subsection{Relative Pose Regression}\label{sec:regression_methods}
\subsectionvspace
Relative pose regression (RPR) networks learn to predict metric relative poses in a forward pass.
We implement a baseline architecture following best practices reported in the literature \cite{zhou2020essnet,WinkelbauerICRA21,zhou2019continuity} -- see Fig.~\ref{fig:network_architecture}, and the supplement for more details.
In the following, we focus on the different output parameterizations and leave a discussion about losses and other design choices to the supplement.

RPR networks often parameterize rotations as \textbf{quaternions}~\cite{en2018rpnet,melekhov2017relative,WinkelbauerICRA21} (denoted as $R(q)$). \cite{zhou2019continuity} argues that a \textbf{6D parameterization} of rotation avoids discontinuities of other representations: 
the network predicts two 3D vectors and creates an orthogonal basis through a partial Gram-Schmidt process (denoted as $R(6D)$). 
Finally, for rotation, we experiment with \textbf{Discrete Euler angles} \cite{cai2021extreme}, denoted as $R(\alpha,\beta,\gamma)$. 
Following \cite{cai2021extreme}, we use 360 discrete values for yaw and roll, and 180 discrete values for the pitch angle.
For the translation vector we investigate three parameterization options: predicting the 
\textbf{scaled translation} (denoted as $t$), predicting a \textbf{scale and unitary translation} separately (denoted as $s\cdot \hat{t}$), and \textbf{scale and discretized unitary translation}.
For the latter we predict translation in spherical coordinates $\phi, \theta$ with quantized bins of 1deg as well as a 1D scale (denoted as $s\cdot \hat{t}(\phi, \theta)$).
As an alternative which model rotation and translation jointly,
we adapt the method of~\cite{suwajanakorn2018discovery} which predicts 3D-3D correspondences for predefined keypoints of specific object classes.
Here, we let the network predict \textbf{three 3D-3D correspondences} (denoted as $[3D-3D]$). 
We compute the transformation that aligns these two sets of point triplets using Procrustes, which gives the relative rotation and translation between the two images. 
The models are trained end-to-end until convergence by supervising the output pose with the ground truth relative pose. We experimented with different loss functions and weighting between rotation and translation losses. For details, see supplemental material.

\section{Map-free Relocalization Datasets}\label{sec:dataset}
\sectionvspace

In this section, we first discuss popular relocalization datasets and their limitations for map-free relocalization. 
Then, we introduce the Niantic map-free relocalization dataset which was collected specifically for the task. Finally, we define evaluation metrics used to benchmark baseline methods.

\subsection{Existing Relocalization Datasets}
\subsectionvspace

One of the most commonly used datasets for visual relocalization is 7Scenes~\cite{shotton2013scene}, consisting of seven small rooms scanned with KinectFusion~\cite{izadi2011kinectfusion}.
12Scenes~\cite{valentin2016learning} provides a few more, and slightly larger environments, while RIO10~\cite{wald2020beyond} provides 10 scenes focusing on condition changes between mapping and query images.
For outdoor relocalization, Cambridge Landmarks~\cite{kendall2015posenet} and Aachen Day-Night \cite{sattler2018benchmarking}, both consisting of large SfM reconstructions, are popular choices. 

We find existing datasets poorly suited to benchmark map-free relocalization.
Firstly, their scenes are not well captured by a single image which holds true for both indoor rooms and large-scale outdoor reconstructions. 
Secondly, the variability across scenes is extremely limited, with 1-12 distinct scenes in each single dataset.
For comparison, our proposed dataset captures 655 distinct outdoor places of interest with 130 reserved for testing alone.
Despite these issues, we have adapted the 7Scenes dataset to our map-free relocalization task.

Regarding relative pose estimation, ScanNet~\cite{dai2017scannet} and MegaDepth~\cite{li2018megadepth} have become popular test beds, \eg for learning-based 2D correspondence methods such as SuperGlue~\cite{sarlin2020superglue} and LoFTR~\cite{sun2021loftr}. 
However, both datasets do not feature distinctive mapping and query sequences as basis for a relocalization benchmark. 
Furthermore, MegaDepth camera poses do not have metric scale.
In our experiments, we use ScanNet~\cite{dai2017scannet} as a training set for scene-agnostic relocalization methods to be tested on 7Scenes. In the supplemental material, we also provide ablation studies on metric relative pose accuracy on ScanNet. 

\subsection{Niantic Map-free Relocalization Dataset}
\begin{figure}[t]
  \centering
  \includegraphics[width=1.0\linewidth, trim={0.15in 10.5in 0.1in 0.3in},clip]{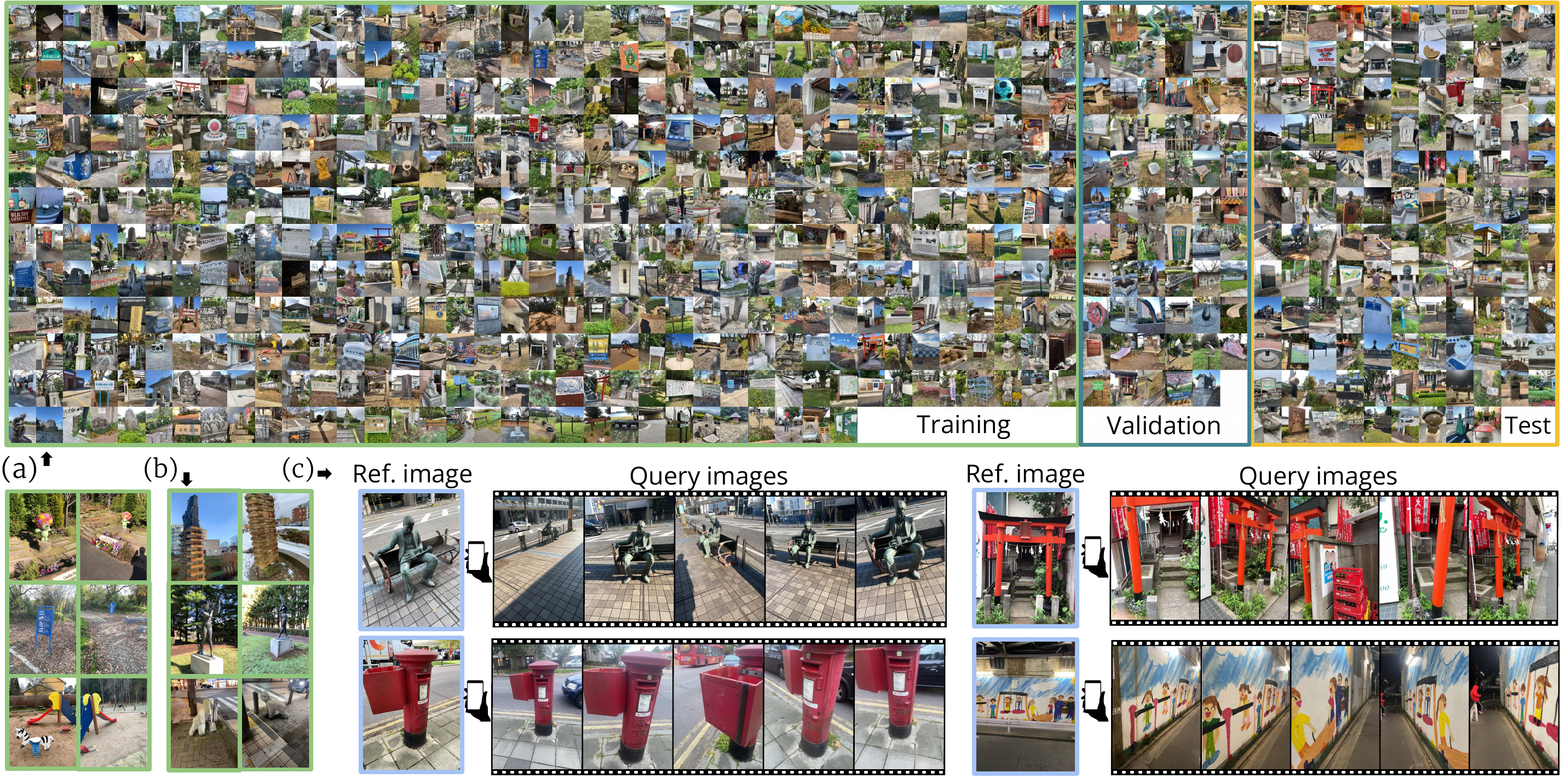}
   \caption{Niantic map-free relocalisation dataset. (a) Dataset overview. Training (460 scenes), validation (65) and test (130) thumbnails. Better seen in color and magnified in electronic format. (b) Examples of training pairs sampled from training scenes. (c) Reference frame (enclosed in \textcolor{carolinablue}{blue}) and an example of query images. Query sequences have been sampled at relative temporal frames: 0\%, 25\%, 50\%, 75\% and 100\% of the sequence duration.}
   \label{fig:dataset1}
   \captionvspace
\end{figure}

\subsectionvspace
We introduce a new dataset for development and evaluation of map-free relocalization. The dataset consists of 655 outdoor scenes, each containing a small `place of interest' such as a sculpture, sign, mural, etc, such that the place can be well-captured by a single image. Scenes of the dataset are shown in Fig.~\ref{fig:dataset1}.

The scenes are split into 460 training scenes, 65 validation scenes, and 130 test scenes.
Each training scene has two sequences of images, corresponding to two different scans of the scene. We provide the absolute pose of each training image, which allows determining the relative pose between any pair of training images. We also provide overlap scores between any pair of images (intra- and inter-sequence), which can be used to sample training pairs.
For validation and test scenes, we provide a single reference image obtained from one scan and a sequence of query images and absolute poses from a different scan.
Camera intrinsics are provided for all images in the dataset.

\begin{figure}[th]
  \centering
  \includegraphics[width=1.0\linewidth, trim={0.03in 9.57in 2.55in 0in},clip]{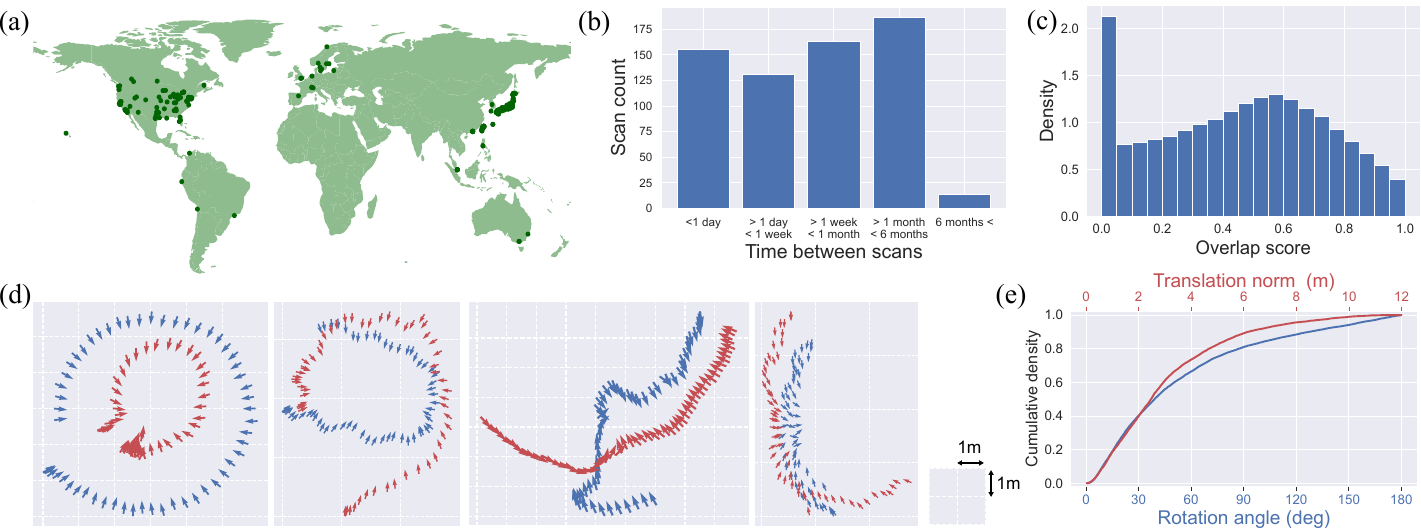}
   \caption{Niantic map-free dataset statistics. (a) Geographic location of scans. (b) Time elapsed between two different scans from the same scene. (c) Visual overlap between training frames estimated using co-visible SfM points, inspired by \cite{sarlin2020superglue}. (d) Sample of different dataset trajectories seen from above. Each plot represents one scene and shows two different trajectories corresponding to two different scans, one in each color. The direction of the arrows represent the camera viewing direction. Each trajectory has been subsampled for visualization. (e) Relative pose distribution between reference image and query images in the test set.}
   \label{fig:dataset_stats}
   \captionvspace
\end{figure}

The Niantic map-free dataset was crowdsourced from members of the public who scanned places of interest using their mobile phones.
Each scan contains video frames, intrinsics and (metric) poses estimated by ARKit (iOS)~\cite{arkit} or ARCore (Android)~\cite{arcore} frameworks and respective underlying implementations of Visual-Inertial Odometry and Visual-Inertial SLAM.
We use automatic anonymization software to detect and blur faces and car license plates in frames.
Scans were registered to each other using COLMAP~\cite{schoenberger2016sfm}. First, we bundle adjust the scans individually by initializing from raw ARKit/ARCore poses.
Then, the two 3D reconstructions are merged into a single reconstruction by matching features between scans, robustly aligning the two scans and bundle adjusting all frames jointly.
We then compute a scale factor for each scan, so that the frames of the 3D reconstructions of each scan would (robustly) align to the raw ARKit/ARCore poses.
Finally, the 3D reconstruction is rescaled using the average scale factor of the two scans.
Further details are provided in supplemental material.
Poses obtained via SfM constitute only a \emph{pseudo} ground truth, and estimating their uncertainty bounds has recently been identified as an open problem in relocalization research \cite{brachmann2021limits}.
However, as we will discuss below, given the challenging nature of map-free relocalization, we evaluate at much coarser error threshold than standard relocalization works.
Thus, we expect our results to be less susceptible to inaccuracies in SfM pose optimization.

The places of interest in the Niantic map-free dataset are drawn from a wide variety of locations around the world and captured by a large number of people. 
This leads to a number of interesting challenges, such as variations in the capture time, illumination, weather, season, and cameras, and even the geometry of the scene;
and variations in the amount of overlap between the scans.
Fig.~\ref{fig:dataset_stats} summarizes these variations.

\subsection{Evaluation Protocol}
\subsectionvspace
Our evaluation protocol consists of rotation, translation and reprojection errors computed using ground truth and estimated relative poses that are predicted for each query and reference image pair.
Given estimated $(R, t)$ and ground truth $(R_\text{gt}, t_\text{gt})$ poses, we compute the rotation error as the angle (in degrees) between predicted and ground truth rotations, $\measuredangle{(R,R_\text{gt})}$.
We measure the translation error as the Euclidean distance between predicted $c$ and ground truth $c_\text{gt}$ camera centers in world coordinate space, where $c=-R^T t$.

Our proposed reprojection error provides an intuitive measure of AR content misalignment.
We were inspired by the Dense Correspondence Reprojection Error (DCRE)~\cite{wald2020beyond} which measures the average Euclidean distance between corresponding original pixel positions and reprojected pixel positions obtained via back-projecting depth maps. 
As our dataset does not contain depth maps we cannot compute the DCRE. 
Hence, we propose a Virtual Correspondence Reprojection Error (VCRE): ground truth and estimated transformations are used to project virtual 3D points, located in the query camera's local coordinate system. 
VCRE is the average Euclidean distance of the reprojection errors:
\begin{equation}
    \text{VCRE} = \frac{1}{|\mathcal{V}|} \sum_{\textbf{v} \in \mathcal{V}} \left\Vert \pi(\textbf{v}) - \pi(T T_\text{gt}^{-1} \textbf{v}) \right\Vert_2~~\text{with}~~T = [R|t],
\end{equation}
where $\pi$ is the image projection function, and $\mathcal{V}$ is a set of 3D points in camera space representing virtual objects.
For convenience of notation, we assume all entities are in homogeneous coordinates.
To simulate an arbitrary placement of AR content, we use a 3D grid of points for $\mathcal{V}$ (4 in height, 7 in width, 7 in depth) with equal spacing of 30 cm and with an offset of 1.8m along the camera axis. 
See supplemental material for a video visualisation, and an ablation showing that DCRE and VCRE are well-aligned.
In standard relocalization, best methods achieve a DCRE below a few pixels \cite{brachmann2021limits}.
However, map-free relocalization is more challenging, relying on learned heuristics to resolve the scale ambiguity. 
Thus, we apply more generous VCRE thresholds for accepting a pose, namely 5\% and 10\% of the image diagonal.
While a 10\% offset means a noticeable displacement of AR content, we argue that it can still yield an acceptable AR experience.

Our evaluation protocol also considers the confidence of pose estimates. 
Confidence enables the relocalization system to flag and potentially reject unreliable predictions.
This is a crucial capability for a map-free relocalization system to be practical since a user might record query images without any visual overlap with the reference frame.
A confidence can be estimated as the number of inlier correspondences in feature matching baselines.
Given a confidence threshold, we can compute the ratio of query images with confidence greater-or-equal to the threshold, \ie the ratio of confident estimates or the ratio of non-rejected samples.
Similarly, we compute the precision as the ratio of non-rejected query images for which the pose error (translation, rotation) or the reprojection error is acceptable (below a given threshold).
Each confidence threshold provides a different trade-off between the number of images with an estimate and their precision. 
Models that are incapable of estimating a confidence will have a flat precision curve.

\section{Experiments}
\sectionvspace

We first report experiments on the 7Scenes~\cite{shotton2013scene} dataset, demonstrating that our baselines are competitive with the state of the art when a large number of mapping images is available. 
We also show that as the number of mapping images reduces, map-free suitable methods degrade more gracefully than traditional approaches. 
Additional relative pose estimation experiments on ScanNet~\cite{dai2017scannet} are reported in the supplement, to allow comparison of our baselines against previous methods.
Finally, we report performance on the new Niantic map-free relocalization dataset and identify areas for improvement.

\begin{figure}[t]
  \centering
  \includegraphics[width=\linewidth]{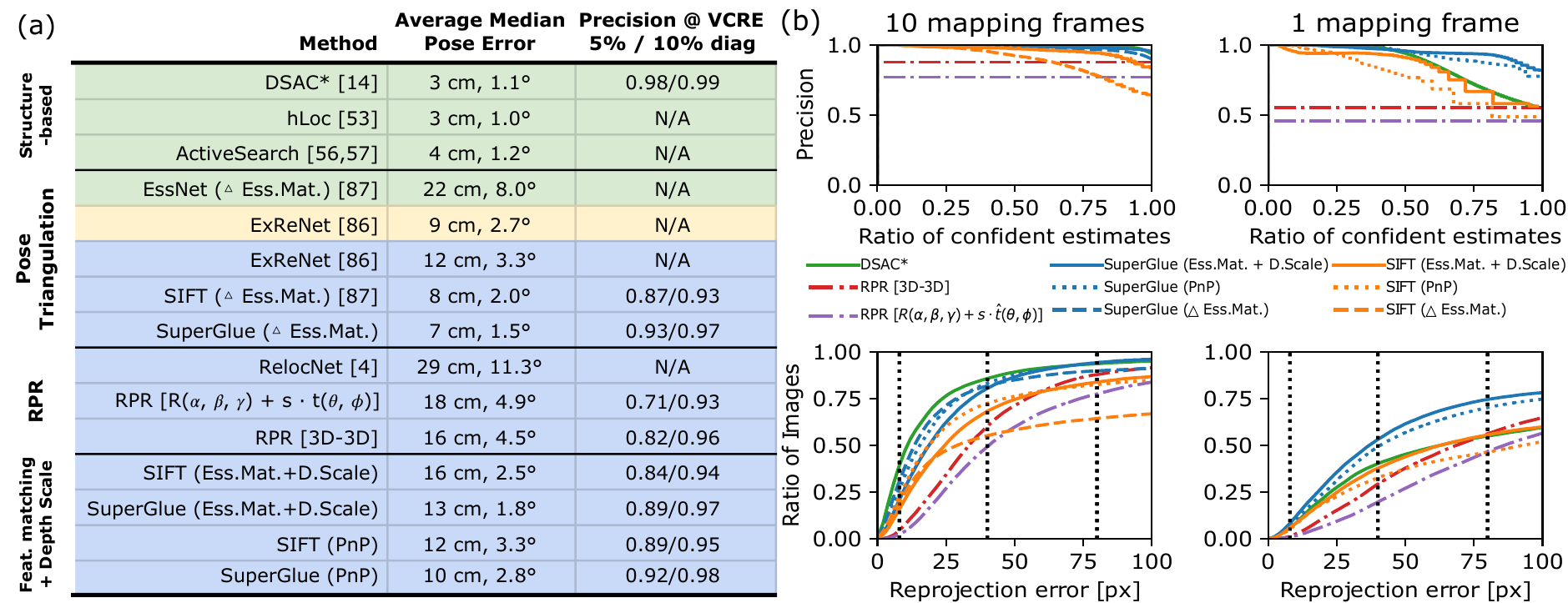}
  \caption{7Scenes results. %
  (a) Using all mapping frames. Dataset-specific (7Scenes) methods in \textcolor{asparagus}{green}, trained on SUNCG \cite{song2016ssc} in \textcolor{goldenrod}{yellow}, and trained on Scannet \cite{dai2017scannet} in \textcolor{carolinablue}{blue}. %
  (b) 10 and 1 mapping frame scenarios: precision curves (top), cumulative density of reprojection error (bottom). %
  Dashed vertical lines indicate 1\%, 5\% and 10\% of the image diagonal, correspondingly 8px, 40px and 80px. %
  }
  \label{fig:7scenes-results}
  \captionvspace
\end{figure}

\subsection{7Scenes}
\subsectionvspace

First, we compare methods described in Sec.~\ref{sec:feature_matching_depth} and Sec.~\ref{sec:regression_methods} against traditional methods when all mapping frames are available. Fig.~\ref{fig:7scenes-results}a shows impressive scores of structure-based DSAC*~\cite{Brachmann2021dsacstar} (trained with depth from PlaneRCNN~\cite{liu2019planercnn}), hLoc~\cite{sarlin2019coarse} and ActiveSearch~\cite{sattler2012improving,Sattler2017AS}.
When 5 reference frames can be retrieved for each query using DenseVLAD~\cite{torii2015denseVLAD} (following \cite{zhou2020essnet}), triangulation-based relative pose methods are competitive with structure-based methods, especially in average median rotation error.
See results for EssNet\cite{zhou2020essnet}, ExReNet\cite{zhang2006image} and our feature matching and triangulation baselines, denoted by $\bigtriangleup$.

Closer to map-free relocalization, if for each query frame, we retrieve a single reference frame from the set of mapping images, the accuracy of metric relative pose estimation becomes more important, see the sections for relative pose regression (\emph{RPR}) and \emph{Feature Matching+D.Scale} in Fig.~\ref{fig:7scenes-results}a. 
Unsurprisingly, methods in both families slightly degrade in performance, with Feature Matching + D.Scale methods beating RPR methods. 
However, all baselines remain competitive, despite depth~\cite{liu2019planercnn} and RPR networks being trained on ScanNet~\cite{dai2017scannet} and evaluated on 7Scenes.
High scores for all methods in Fig.~\ref{fig:7scenes-results}a are partially explainable by the power of image retrieval and good coverage of the scene. 

In map-free relocalization, the query and reference images could be far from each other. 
Thus, we evaluate the baselines on heavily sparsified maps, where metric relative pose accuracy is more important. 
We find the $K$ most representative reference images of each scene by $K$-means clustering over DenseVLAD~\cite{torii2015denseVLAD} descriptors of the mapping sequence. 
In Fig.~\ref{fig:7scenes-results}b we show results for $K=10$ and $K=1$, where $K=1$ corresponds to map-free relocalization. 
We show precision curves using a pose acceptance threshold of VCRE $<$ 10\% of the image diagonal (\ie 80px). 
We also plot the cumulative density of the VCRE.
Unsurprisingly, pose triangulation methods fare well even when $K=10$ but cannot provide estimates when $K=1$. 
For $K=1$, Feature Matching+D.Scale outperforms the competition. 
Specifically, SuperGlue (Ess.Mat.+ D.Scale) recovers more than 50\% of query images with a reprojection error below 40px. 

DSAC* remains competitive in sparse regimes, but it requires training per scene, while the other baselines were trained on ScanNet. 
Both ScanNet and 7Scenes show very similar indoor scenes. 
Yet, single-image depth prediction seems to generalize better across datasets compared to RPR methods, as Feature Matching+D.Scale methods outperform RPR baselines both with $K=10$ and $K=1$ scenarios. 
RPR methods perform relatively well for larger accuracy thresholds but they perform poorly in terms of precision curves due to their lack of estimated confidence.
Further details on all baselines, qualitative results and additional ablation studies can be found in the supplement.

\subsection{Niantic map-free relocalization dataset}
\subsectionvspace

\begin{figure}[tb]
  \centering
  \includegraphics[width=\linewidth]{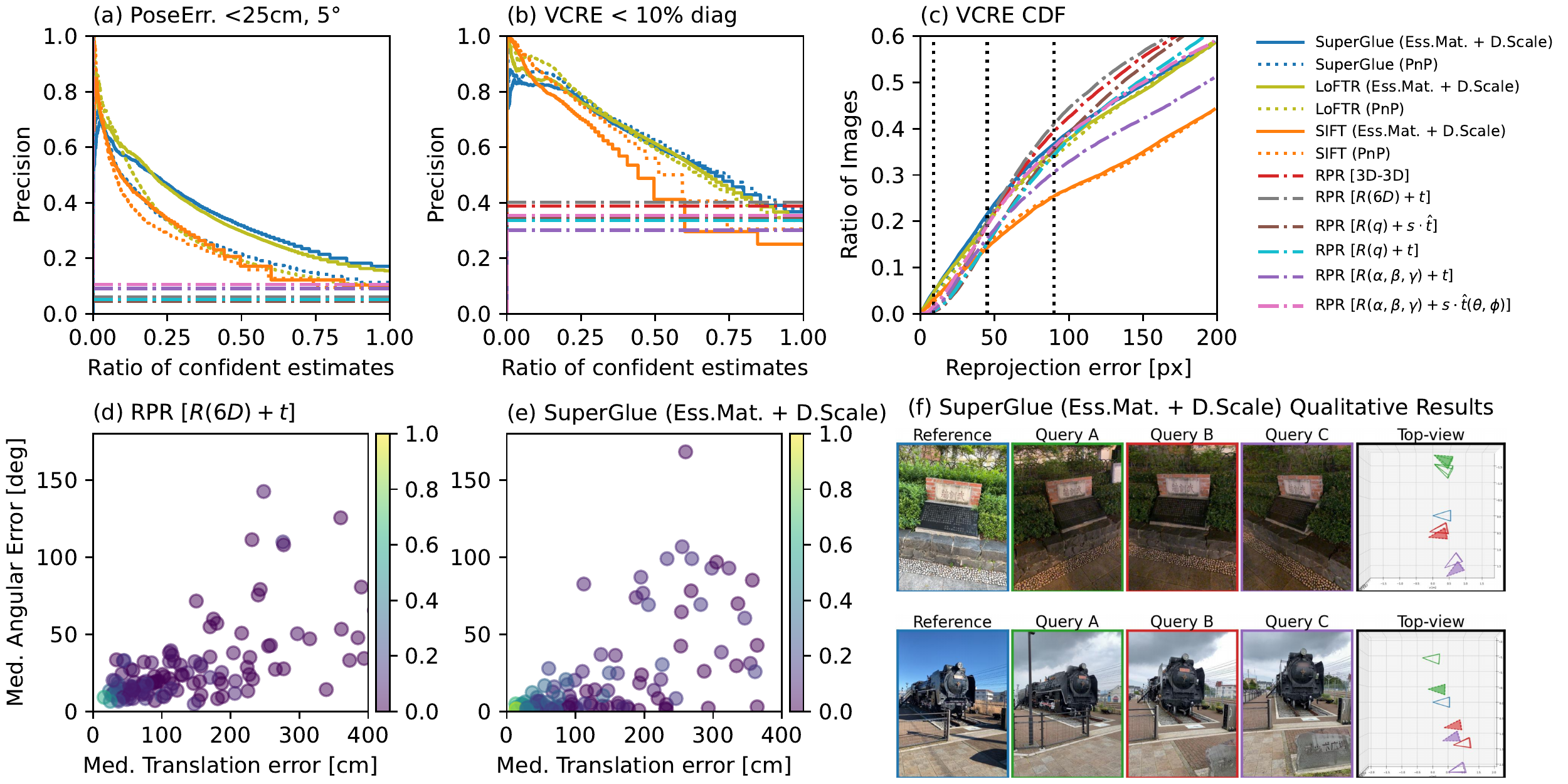}
  \caption{Our dataset results. (a,b) Precision plots using pose error (a) and VCRE (b) thresholds. (c) VCRE CDF, vertical lines indicate 1, 5 and 10\% of the image diagonal, corresp. 9px, 45px and 90px.%
  (d,e) Scatter plot of median angular vs translation error for each scene, estimated using RPR $[R(6D)+t]$ (d) and SG \cite{sarlin2020superglue} Ess.Mat.+D.scale (e). Each point represents a scene, and the colormap shows precision for pose error threshold $25$cm, $5^\circ$. (f) Qualitative results: the reference frame and three queries are shown for two scenes. The top view shows the ground truth (solid line, no fill) and estimated poses (dashed line, filled).}
  \label{fig:ourdataset-results}
  \captionvspace
\end{figure}

Fig.~\ref{fig:ourdataset-results} shows our main results on the Niantic map-free dataset.
As seen in Fig~\ref{fig:ourdataset-results} a, b and c, this dataset is much more challenging than 7Scenes for all  methods. 
This is due to multiple factors: low overlap between query and reference images; the quality of feature matching is affected by variations in lighting, season, etc; and the use of single-image depth prediction networks trained on KITTI for non-driving outdoor images.

In Fig.~\ref{fig:ourdataset-results}d and \ref{fig:ourdataset-results}e we show results of the best methods in each family of baselines: RPR with 6D rotation and scaled translation parameterization and SuperGlue (Ess.Mat.+D.Scale).
SuperGlue (Ess.Mat.+D.Scale) in Fig.~\ref{fig:ourdataset-results}e reports a median angular rotation below $10^\circ$ for a large number of scenes. 
In these cases, the high variance of the median translation error is partly due to the variance of depth estimates. Further improvement of depth prediction methods in outdoor scenes should improve the metric accuracy of the translation error.
Qualitative examples in Fig~\ref{fig:ourdataset-results}f shows where depth improvements could produce better results: both the angular pose and the absolute scale in the first row are accurate, while the second row has good angular pose and bad absolute scale.

The RPR method in Fig.~\ref{fig:ourdataset-results}d exhibits a different behavior: the average angular error is lower than for Feature Matching+D.Scale baselines, yet it rarely achieves high accuracy. 
This is also evident in Fig.~\ref{fig:ourdataset-results} c, where Feature Matching+D.Scale methods outperform RPR methods for stricter thresholds, but degrade for broader thresholds. 
Indeed, when the geometric optimization fails due to poor feature matches, the estimated scaleless pose can be arbitrarily far from the ground truth.
In contrast, RPR methods fail more gracefully due to adhering to the learned distribution of relative poses. 
For example, in Fig.~\ref{fig:ourdataset-results}c allowing for a coarser VCRE threshold of 10\% of the image diagonal, the $[3D-3D]$ and $[R(6D)+t]$ variants overtake all methods, including feature matching-based methods.
Hence, RPR methods can be more accurate than feature matching at broad thresholds, but they offer lower precision in VCRE at practical thresholds.

RPR methods currently do not predict a confidence which prevents detecting spurious pose estimates, \eg when there
is no visual overlap between images, as illustrated in the supplement.
Although feature matching methods can estimate the confidence based on the number of inliers, the precision curves in Fig.~\ref{fig:ourdataset-results}a show that these confidences are not always reliable. 
Further research in modeling confidence of both families of methods could allow to combine their advantages.

\section{Conclusion and Future Work}
\sectionvspace

We have proposed map-free relocalization, a new relocalization task. 
Through extensive experiments we demonstrate how existing methods for single-image depth prediction and relative pose regression can be used to address the task with some success.
Our results suggest some directions for future research: improve the scale estimates by improving depth estimation in outdoor scenes;
improve the accuracy of metric RPR methods; and derive a confidence for their estimates.

To facilitate further research, we have presented the Niantic map-free relocalization dataset and benchmark with a large number of diverse places of interest. We define an evaluation protocol to closely match AR use cases, and make the dataset and an evaluation service publicly available.

As methods for this task improve, we hope to evaluate at stricter pose error thresholds corresponding to visually more pleasing results.
A version of map-free relocalization could use a burst of query frames rather than a single query frame to match some practical scenarios. Our dataset is already suitable for this variant of the task, so we hope to explore baselines for it in the future.

\noindent
\textbf{Acknowledgements}
We thank Pooja Srinivas, Amy Duxbury, Camille Fran\c{c}ois, Brian McClendon and their teams, for help with validating and anonymizing the dataset and Galen Han for his help in building the bundle adjustment pipeline. We also thank players of Niantic games for scanning places of interests.

\clearpage
%
%
\bibliographystyle{splncs04}
\bibliography{egbib}

\end{document}